\documentclass[sigconf]{acmart}
\usepackage{amsmath,amsfonts}
\usepackage{algorithmic}
\usepackage{array}
\usepackage{ragged2e}
\usepackage{ multirow, tabularx}
\AtBeginDocument{%
  \providecommand\BibTeX{{%
    \normalfont B\kern-0.5em{\scshape i\kern-0.25em b}\kern-0.8em\TeX}}}

\copyrightyear{2023}
\acmYear{2023}
\setcopyright{acmlicensed}
\acmConference[MM '23] {Proceedings of the 31st ACM International Conference on Multimedia}{October 29--November 3, 2023}{Ottawa, ON, Canada.}
\acmBooktitle{Proceedings of the 31st ACM International Conference on Multimedia (MM '23), October 29--November 3, 2023, Ottawa, ON, Canada}
\acmPrice{15.00}
\acmISBN{979-8-4007-0108-5/23/10}
\acmDOI{10.1145/3581783.3611878}

\begin{document}

\title{A Reinforcement Learning-Based Automatic Video Editing Method Using Pre-trained Vision-Language Model}

\author{Panwen Hu}
\orcid{0000-0001-6183-6598}
\affiliation{%
  \institution{SSE, The Chinese University \\
  of Hong Kong, Shenzhen}
  \country{China}
  \postcode{518172}
}
\email{panwenhu@link.cuhk.edu.cn}

\author{Nan Xiao}
\orcid{0009-0003-0886-9982}
\affiliation{%
  \institution{SSE, The Chinese University \\
  of Hong Kong, Shenzhen}
  \country{China}
  \postcode{518172}
}
\email{nanxiao@link.cuhk.edu.cn}

\author{Feifei Li}
\orcid{0009-0006-8098-0265}
\affiliation{%
  \institution{SSE, The Chinese University \\
  of Hong Kong, Shenzhen}
  \country{China}
  \postcode{518172}
}
\email{feifeili1@link.cuhk.edu.cn}

\author{Yongquan Chen}
\affiliation{%
   \institution{AIRS, The Chinese University \\
   of Hong Kong, Shenzhen}
  \country{China}
  \postcode{518172}}
\email{yqchen@cuhk.edu.cn}

\author{Rui Huang}
\authornote{Rui Huang is the corresponding author. This work was partially supported by Shenzhen Science and Technology Program (JCYJ20220818103006012, JCYJ20210324115604012, ZDSYS20211021111415025, ZDSYS20220606100601002), and Shenzhen Institute of Artificial Intelligence and Robotics for Society (AIRS).}
\affiliation{%
   \institution{SSE, The Chinese University \\
   of Hong Kong, Shenzhen}
  \country{China}
  \postcode{518172}}
\email{ruihuang@cuhk.edu.cn}


\begin{abstract}
  In this era of videos, automatic video editing techniques attract more and more attention from industry and academia since they can reduce workloads and lower the requirements for human editors. Existing automatic editing systems are mainly scene- or event-specific, e.g., soccer game broadcasting, yet the automatic systems for general editing, e.g., movie or vlog editing which covers various scenes and events, were rarely studied before, and converting the event-driven editing method to a general scene is nontrivial. In this paper, we propose a two-stage scheme for general editing. Firstly, unlike previous works that extract scene-specific features, we leverage the pre-trained Vision-Language Model (VLM) to extract the editing-relevant representations as editing context. Moreover, to close the gap between the professional-looking videos and the automatic productions generated with simple guidelines, we propose a Reinforcement Learning (RL)-based editing framework to formulate the editing problem and train the virtual editor to make better sequential editing decisions. Finally, we evaluate the proposed method on a more general editing task with a real movie dataset. Experimental results demonstrate the effectiveness and benefits of the proposed context representation and the learning ability of our RL-based editing framework.

%
\end{abstract}




\maketitle

\section{Introduction}

With the development of multimedia technology, video has become a prime  medium for conveying information \cite{Zhang2022}, thus creating high-quality videos becomes more and more crucial. Typically, video editing involves the shot selection problem and requires the editors to select and order the shots from a vast footage gallery. However, editing video or live multi-camera directing requires a multitude of skills and domain expertise. Even for professional editors/directors, editing and directing video streams are still challenging and labor-intensive processes \cite{Saini2018}. 

\begin{figure}[htbp]
  \includegraphics[scale=0.8]{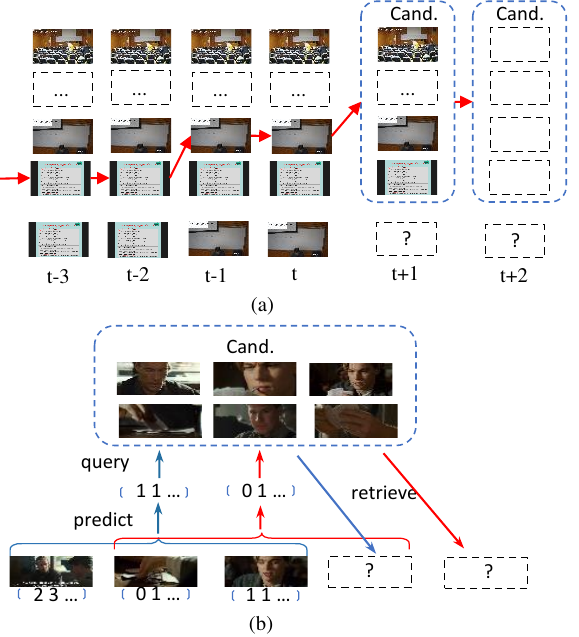}
  \vspace{-6pt}
  \caption{Definitions of different editing tasks. (a) is event-driven editing; (b) is attribute prediction-based editing.}
  \label{fig:problem definition}
\end{figure}

A few automatic editing systems \cite{Shrestha2010, Saini2012, Wu2015, Bano2016} and directing systems \cite{Ariki2006, Wang2008, Kaiser2012} have been proposed for event-driven scenes. The editing process is illustrated in Fig.\ref{fig:problem definition}(a), where the virtual editor selects a view based on the scene events at each time point. In previous works, different view selection strategies have been developed, such as the Finite State Machine (FSM) model \cite{Wang2007}, trellis graph model \cite{Arev2014}, integer programming model \cite{Pan2021}, Markov Decision Problem (MDP) model\cite{Hu2021}, etc. However, there are still some drawbacks to these systems. On the one hand, these event-driven methods are scene-specific, and the features and selection algorithms are designed based on the demands of the scenes. This fact limits them to adapt to a more general setting ( Fig.\ref{fig:problem definition}(b)) that focuses on retrieving the shots of various scenes sequentially from a candidate list based on the given shot context. On the other hand, although these systems are developed to mimic human editors using various scenario-related hand-crafted features and empirical rules \cite{Germeys2007, Gottlieb2013}, the quality of their productions heavily depends on empirical parameter tuning. There is still an aesthetic gap between automatic productions and professional-looking videos due to the difficulties in transforming empirical practices into computable formulas. To solve the scene diversity problem for the general editing in Fig.\ref{fig:problem definition}(b),  this paper first defines a proxy editing task and explores a general editing representation. Furthermore, we posit that the view selection (Fig.\ref{fig:problem definition} (a)) can be considered as a particular variant of shot selection (Fig.\ref{fig:problem definition}(b)) with a strict timeline constraint. Thereby, we propose a general learning-based editing framework to mitigate the aesthetic gap problem for both two settings.



We propose a new editing paradigm to accommodate diverse scenes for general editing, e.g., movie editing, as shown in Fig.\ref{fig:problem definition}(b). Specifically, the editing task is to predict the multi-dimensional attributes of subsequent shots, since shot attributes are common in different scenes and determine the types of shots. Thereby, the editor is able to retrieve or create the shots from the raw footage based on the predicted attributes. Furthermore, the gap between professional productions and automated productions also exists in the automatic cinematography area, which focuses on automatically placing and moving the cameras to capture well-composed photos or videos \cite{Heck2007, Chen2016}. To mitigate this problem, some recent works \cite{Gschwindt2019, Huang2019, Jiang2020} start studying the methods that learn a camera controller from expert behaviors or human productions. Inspired by these works, this paper will explore a learning-based strategy to learn the editing pattern styles from professional demonstrations.  


Learning the editing pattern from demonstration videos of the general scene is nontrivial. Like the editing process of existing event-driven editing systems \cite{Saini2012, Arev2014, Wu2015}, the learning-based editing undergoes two stages, i.e., the representation extraction of context shots and the training of the virtual editor. As the representation for general editing is rarely studied before, it is still challenging to distill the context information for editing general scenes. Recently, pre-trained Vision-Language Model (VLM) \cite{Radford2021} has successfully assisted various vision tasks \cite{Minderer2022, Xu2022} due to its great generalization and adaption abilities. Motivated by this, we propose to leverage VLM's extraordinary zero-shot recognition capability to extract the attribute distributions of context shots as the context representations. Afterward, a transformer-based neural network is developed to encode the temporal relationship of context representations for the prediction task. Relevant experiments are conducted in Sec.\ref{sec:experiments}.

The methodology of training a virtual editor also plays a crucial role. Previous methods have attempted to train a Long Shot-Term Memory (LSTM) model \cite{Chen2018, Wu2018} as a virtual director with manual annotations in a supervised manner, while the virtual director/editor trained in a conventional supervised way cannot produce optimal sequential editing results. To this end, we propose an RL-based editing framework to optimize the virtual editor for making better sequential editing decisions. The details are discussed in Sec.\ref{sec:framwork}.

To investigate the style/pattern learning capability and the generalization of the proposed framework,  we apply it to the task of learning personal preferences in an event-driven scene, i.e., the automatic multi-camera lecture broadcasting \cite{Liu2001, Rui2004, Wang2007} scene, in Sec.\ref{sec:experiments}. Finally, we design a set of metrics to quantitatively evaluate the effectiveness of the proposed representations and the editing framework on the AVE dataset \cite{Argaw2022}. To the best of our knowledge, this is the first work to explore the knowledge of VLM for general editing purposes. It is also the first time to study the RL technique for automatic editing/directing tasks. In summary, we have made the following contributions:
\begin{itemize}
  \item We attempt to leverage the knowledge of pre-trained VLM to represent the context shots for general editing purposes.
  \item We propose an RL-based editing framework to mitigate the video quality gap and optimize the sequential decisions.
  \item We develop a set of metrics and conduct experiments to validate the benefits of the proposed components. We further show an application of our framework to a real-world broadcasting scene to verify its learning capability.
\end{itemize}

\section{Related Work}
\label{sec:related work}
\subsection{Automatic Video Editing}
Automatic video editing/directing has attracted much attention recently \cite{Saini2018,Soe2021, Ronfard2021, Zhang2022}. As a core part of video editing, various camera view selection algorithms have also been proposed for different applications. For example, Wang et al.\cite{Wang2008, Yang2019} define a set of selection rules in their system to select the camera for broadcasting soccer games, mimicking an experienced broadcaster. Some other systems \cite{Shrestha2010, Bano2016} adopted a simple yet feasible greedy selection strategy for editing user-generated videos, in which the cameras are selected to maximize the pre-defined scores. For the scene of online lecture broadcasting, the FSM has been employed to model the view-switching process in previous works \cite{Liu2001, Rui2004, Wang2007}. Moreover, Daniyal et al.\cite{Daniyal2011} adopted a Partially Observable Markov Decision Process (POMDP) to model the view selection in editing basketball videos. There are also some studies \cite{Jiang2008, Wang2014, Wu2015, Pan2021} that formulate the camera selection as an optimization problem, which can be solved with dynamic programming techniques.

Nevertheless, the productions from these heuristic methods cannot always fulfill audiences' preferences or professional styles. These methods are developed upon a set of scene-specific features, e.g., visibility of objects-of-interest \cite{Chen2013, Wang2014}, the size of the object \cite{Wang2014}, transition constraints \cite{Chen2013, Wang2014, Bano2016}, and stability \cite{Wang2008, Bano2016}, etc. Although they are expected to represent professional editing/directing practices precisely, there is a gap in different preferences and a gap in the transformation from a library of features to professional editing elements. To this end, another line of works studies the data-driven camera selection strategies and trains a regressor \cite{Chen2018} to rank the camera's importance. Yet these methods require large numbers of real-valued labels(visual importance). In addition, these data-driven methods, and the above heuristic methods, are limited to particular working scenes. Therefore, this paper explores a more general editing representation and proposes an RL-based editing framework for editing videos of general scenes. 


\subsection{Learning-Based Video Cinematography}
Video cinematography is also a crucial part of the art of video production. Automated cinematography also encounters a similar problem: the productions from rule-based cinematography are far from satisfactory \cite{Yu2022}. Therefore, research in cinematography started to explore learning-based methods. In drone cinematography, Gschwindt et al.\cite{Gschwindt2019} proposed a system trained under a deep RL framework to choose the shot type that maximizes a reward based on a handcrafted aesthetic metric. This work is further improved by Bonatti et al.\cite{Bonatti2020} to solve the occlusion avoidance problem and optimize the trajectory. Another line of works \cite{Huang2019, Huang2021} for drone cinematography train a prediction network to estimate the next location and shot angle of the drone, given the past locations and the pose of the character. The goal of these works is to make the style of autonomous videos approach that of trained videos. Besides drone cinematography, some other works also have studied autonomous cinematography for filmmaking \cite{Jiang2020, Yu2022} or photograph \cite{AlZayer2021} using deep RL. The reward functions promoting the agent training are derived from the aesthetics measurements on frame content or a trained model with aesthetic analysis dataset \cite{Murray2012}. However, it is a non-trivial task to apply the learning algorithms of autonomous cinematography to autonomous editing. The cinematography style or aesthetics reward can be measured on the frame content, but measuring the editing style of the camera sequence is challenging.

\section{The Proposed Method}
\label{sec:framwork}

\begin{figure*}[htbp]
  \includegraphics[scale=0.8]{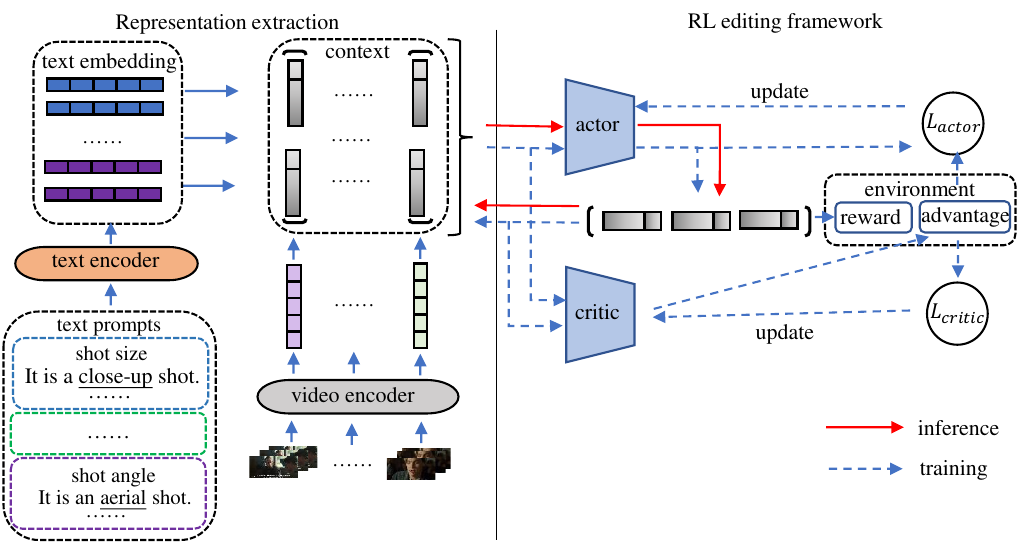}
  \vspace{-6pt}
  \caption{The architecture of the proposed method and its training and testing processes.}
  \label{fig:system}
\end{figure*}


\subsection{Problem Definition}



Although the concept of automatic editing has been proposed for a long time, previous systems define the editing task under their own scenes and have limited generalization ability. Therefore, unifying the editing tasks for general scenes is urgently needed. Recently, Argaw et al. \cite{Argaw2022} publish a large-scale movie dataset, AVE, that covers a wide variety of scenes, and the shots of different scenes share a common set of 8 attributes, including shot size, shot angle, shot type, shot motion, shot location, shot subject, number of people, and sound source. To accommodate general editing scenes, we formulate the editing task as predicting the shot attributes of subsequent shots. Concretely, given the historical context shots $\{S_0, S_1, \dots, S_n\}$, our task aims to predict the attributes $\{A_{n+1}, \dots, A_{n+M}\}$ of subsequent shots $\{S_{n+1}, \dots, S_{n+M}\}$. The attributes $A_i=(a_1, \dots, a_8)$ of each shot is an 8-dimension vector, and each element represents a class of the corresponding attribute. As a result, with the predicted attributes $A_{i}$, users can retrieve a suitable shot to assemble from the available candidate list or create a suitable shot from raw footage.

    
To solve the general editing problem, this paper proposes a two-stage scheme that includes representation extraction and virtual editor training stages, as shown in Fig.\ref{fig:system}. In the first stage, we transfer the cross-modal knowledge of VLM to extract the editing-relevant representation without using any other manual label. In the second stage, our editing framework trains a virtual editor/actor based on the extracted features to make sequential decisions.

\subsection{Context Representation}
\label{sec:context representation}
The features used to present the historical context are crucial in video editing. The virtual editor needs to fully understand the high-level semantics of context before making a decision like a human editor, and a proper context representation helps the agent understand the context better. As previous high-level semantics e.g., the positions and numbers of players \cite{Wang2014}, the event \cite{Wang2007}, for event-driven scenes might be unavailable in general scenes, some works \cite{Chen2018, Argaw2022} use pre-trained action recognition model, e.g., R3D \cite{Tran2018}, to extract the context features. However, these deep features are not interpretable and do not carry the semantics needed for editing since the feature model focuses more on the information for discriminating action categories rather than editing.

The context representation should be general and editing-relevant, but no previous work has studied this field before. Fortunately, we notice that pre-trained VLM, e.g., CLIP \cite{Radford2021}, XCLIP \cite{Ma2022}, has shown its strong generalization abilities in different tasks recently, and its comprehensive knowledge is leveraged to improve traditional close-set methods to an open task domain, e.g., open-vocabulary detection \cite{Minderer2022} and segmentation \cite{Xu2022}. Inspired by this, we attempt to transfer the general knowledge of VLM to the editing domain. In this paper, we choose the XCLIP model, which is trained with masses of paired video-text data covering a variety of domains, as the knowledge source, and perform zero-shot recognition ability to extract the attribute distributions of context shots as representation. Specifically, for each attribute $a_i$, we first construct a set of text prompts using all class names of $a_i$. For example, the set of prompts $P_i$ for shot angle attribute is constructed as "\textit{it is a/an \underline{aerial} shot}", " \textit{it is a/an \underline{overhead} shot}", "\textit{it is a/an \underline{low angle} shot}", etc. These text prompts are fed into the text encoder of XCLIP to calculate the prompt embedding $E_{i}^t \in R^{C_i \times d} $ where $C_i$ denotes the number of classes of $a_i$, and the visual embedding $E_{j}^v \in R^{d}$ of context shot $S_j$ are extracted by the video encoder:
\begin{align*}
  E_{i}^t = TextEncoder(P_i), \quad E_{j}^v = VideoEncoder(S_j) 
\end{align*}
The attribute distribution $p_{j,i} \in R^{C_i}$ of $S_j$ over $a_i$ are obtained by applying softmax function on the similarity $D_{j,i} \in R^{C_i}$ between $E_{i}^t$ and $E_{j}^v$, that is
\begin{align*}
  D_{j,i} = E_{i}^t E_{j}^v, \quad p_{j,i}[k] = \frac{e^{D_{j,i}[k]}}{\sum e^{D_{j,i}[k]}}
\end{align*}

We calculate the distributions over all attributes $\{a_1, \dots, a_8\}$ in the same way for each context shot, and the information of $S_{j}$ is represented as $\bar{A}_{j}=[p_{j,1}|\dots|p_{j,8}]$ where $|$ denotes the concatenation operation. As a result, the information of all context shots is defined as $\{\bar{A}_0, \dots, \bar{A}_{n}\}$.

\subsection{RL-Based Editing Framework}
Another essential stage in the automatic video editing method is shot/view selection. Existing heuristic methods integrate various empirical rules summarized by professional experts in the selection process. Yet the difficulties in precisely transferring these rules into computable formulas may result in a gap between the resultant videos and the professional-look videos. In addition, the users are still required to manually fine-tune the parameters to ensure a high-quality resultant video. Recently, some studies start to explore learning-based methods to broadcast a scene automatically. A main difference between broadcasting/directing and general editing is that general editing may involve sequential decisions, while broadcasting can be treated as one-shot editing only. Therefore, traditional supervised learning methods cannot tackle sequential editing well. To overcome this problem, we propose an RL editing framework to handle sequential editing problems. There are three key elements, i.e., state, action, and reward, to be defined.

\textbf{State} The state $E$ in our framework is the information source to guide the virtual editor to make decisions. It can be arbitrary features extracted from the input streams. For example, for event-driven editing, handcrafted features \cite{Lowe2004, Dalal2005} or high-level semantic features \cite{Wang2015,Hu2021} can be used to define a state. In the movie editing scene, we define a state as the attribute distributions of context shots, i.e., $E_0 = [\bar{A}_0|\dots|\bar{A}_{n}]$. State $E_0$ is fed into the actor network to predict the attribute distribution $\hat{A}_{n+1}$ of $S_{n+1}$, then a subsequent state $E_{1}=[\bar{A}_1|\dots|\bar{A}_{n}|\hat{A}_{n+1}]$ is used to predict the attributes $\hat{A}_{n+2}$ of $S_{n+2}$. To intensify the temporal relationship among the shots, we will additionally add position embedding to the shot representation, just like the word embedding in the transformer model will be assigned with positional encodings.

\textbf{Action} The action $O$ is the editing decision by the virtual editor, and the definition of its space depends on the goal of the task. For example, in the directing scene, the action space could be defined as the indices of cameras to be broadcasted. In the movie editing task, we define an action as an 8-dimensional vector, $O \in \mathbb{R}^{8}$, with each element indicating a class of the attribute.

\textbf{Reward} Reward $r$ is an essential part that encourages the virtual editor to make decisions, and it measures how well the action taken is. Technically, a larger reward means that the action token is better. In our task setting, an action is an 8-dimensional vector, so we set the reward at each step as also an 8-dimensional vector to measure the predicted attributes independently. Let $\hat{A}_i \in \mathbb{R}^8$ and $A_i \in \mathbb{R}^8$ denote the action predicting the attributes of $S_{i}$ and the ground-truth attributes, respectively, we define the reward vector as:
\begin{align*}
\label{eq:reward model}
  r &= R(\hat{A}_i,A_i)\\
    &=[\mathbb{I}(\hat{A}_i[1], A_i[1]), \dots, \mathbb{I}(\hat{A}_i[8], A_i[8])], \text{where} \\
\mathbb{I}(\hat{a},a) &= \begin{cases}
                      1, & \mbox{if } \hat{a}=a \\
                      -1, & \mbox{otherwise}.
                    \end{cases}
\end{align*}

\textbf{Training}  We develop our RL framework with an actor-critic scheme \cite{Sutton1999,Lillicrap2015,Schulman2017} to make framework more effective and general. The goal of training is to find the optimal policy/actor network $\pi$ maximizing the total discounted reward $\mathcal{R}$ in an editing episode. Let $r_t$ and $\gamma$ denote the immediate reward at the t-th step and the discount factor, respectively, the total discounted reward $\mathcal{R}$ and the optimal policy $\pi^{*}$ is defined as:
\begin{align*}
  \mathcal{R} &= \sum_{t=0}^{T} \gamma^{t-1}r_t, t=0,1,\dots,T, \quad \pi^{*} = \arg \max \mathbb{E}(\mathcal{R}|\pi)
\end{align*}

To achieve this goal, two networks, a critic network $V$ and a policy/actor network $\pi$ will be trained. The critic network aims to measure how good the action taken is, and its outputs will be used to indicate the updated direction of the policy network.  Specifically, we build a Multi-Layer Perception (MLP) model as the critic network and it takes as inputs the context representation $E_i = [\bar{A}_i|\dots|\bar{A}_{n+i}]$ and the action $\hat{A}_{n+i+1}$ sampled from $\pi(E_i)$, and returns an 8-dimensional vector indicating the quality score of sequence $\{A_i, \dots, A_{n+i}, \hat{A}_{n+i+1}\}$ and the expected future cumulative discounted reward from state $E_{i+1} = [\bar{A}_{i+1}|\dots|\bar{A}_{n+i}|\hat{A}_{n+i+1}]$. Training critic network requires defining the advantage function, which indicates how much reward is gained by taking current action compared to the average decisions.  The advantage function $\mathcal{A} \in \mathbb{R}^{8}$ at time $t$ is defined as:
\begin{equation}\label{eqn:advatage function}
\begin{aligned}
  \mathcal{A} &= \delta_t + \gamma \delta_{t+1} + \dots + \gamma^{T-t+1}\delta_{T-1} \\
  \delta_t &= r_t + \gamma V(E_{t+1}, \hat{A}_{n+t+1}) - V(E_{t}, \hat{A}_{n+t})
\end{aligned}
\end{equation}
The objective function $L_{critic}$ for training $V$ is to minimize the advantage function, $L_{critic} = \frac{1}{2} ||\mathcal{A}||^2$.

We use a transformer-based architecture as the backbone of the actor network due to its excellent ability to explore the temporal relationships among the tokens. The actor network takes as input the editing context $E_i = \{\bar{A}_i,\dots, \bar{A}_{n+i}\}$ and treats each shot representation as a token embedding, and the final global embedding is passed to eight MLP heads followed by a softmax function to generate the distributions $\hat{A}_{n+i+1}$ of 8 attributes. The update direction of policy network $\pi$ will depend on the advantage function $\mathcal{A}$ as it measures the profits of the actions sampled from $\pi$,  thus the objective for training policy network is defined as follows:
\begin{align}
\label{eqn:actor loss}
  L_{actor} &= \sum - \log \pi_\theta(\hat{A}_{n+t}|E_t)\mathcal{A}
\end{align}
Intuitively, when the sampled action $\hat{A}_{n+t}$ leads to a large positive $\mathcal{A}$, the probability of $\hat{A}_{n+t}$ will be increased, and vice verse.

The overall training process at a step is shown in Fig.\ref{fig:system}. At state $E_i$, an action $\hat{A}_{n+i+1}$ is first sampled from $\pi(E_i)$ and passed to the environment to acquire the reward $r_i$, then a new state $E_{i+1}$ is constructed, and a new action $\hat{A}_{n+i+2}$ is sampled. This process proceeds until the maximum length is reached, and a sequence of $(E_i, \hat{A}_{n+i+1}, r_i)_{i=1:T}$ is obtained. Afterward, $\hat{A}_{n+i+1}$ and $E_i$ are fed into $V$ to compute a score, which will be compared with cumulative discounted reward to calculate the advantage of $\hat{A}_{n+i+1}$. This advantage will be used to update $V$ and $\pi$.

\section{Experiments}
\label{sec:experiments}
In this section, we will first introduce the dataset and the evaluation metrics of the proposed new task. Afterward, we will conduct various experiments to validate the superiority of the proposed context representation and the RL-based editing framework. Finally, we will apply our editing framework to an online lecture broadcasting scene, an event-driven scene, to examine its generalization ability.

\begin{table*}[!htbp]
\centering
\begin{tabular}{c|c|c|c|c|c|c|c|c|c|c|c|c|c}
\hline
 Methods & $Acc_{np}$ & $Acc_{sa}$ & $Acc_{sl}$ & $Acc_{sm}$ & $Acc_{ss}$ & $Acc_{sub}$ & $Acc_{st}$ & $Acc_{sou}$ & $1-Acc$ & $2-Acc$ & reward & $rank1$ & $2-rank1$ \\
\hline
\textbf{Random}   & 12.9 & 18.2 & 33.8 & 16.1 & 17.1 & 11.5 & 12.4 & 16.9 & 0.0 & 0.0 & 2.7 & 18.7 & 4.1 \\
\textbf{MLP}      & 39.1 & 85.2 & 81.4 & 54.2 & 73.7 & 82.3 & 46.4 & 83.2 & 7.7 & 2.5 & 10.9 & 42.4 & 25.3 \\
\textbf{LSTM}     & 39.2 & 85.2 & 81.6 & 54.9 & 73.7 & 82.6 & 46.5 & 83.3 & 7.9 & 2.3 & 11.0 & 42.5 & 25.6\\
\textbf{Trans}    & 40.8 & 85.3 & 81.4 & 55.5 & \textbf{75.0} & 83.7 & 46.1 & 83.8 & 8.4 & 2.4 & 11.0 & 43.7 & 24.7 \\
\textbf{RL(Ours)} & \textbf{52.3} & \textbf{86.3} & \textbf{89.1} & \textbf{59.9} & 74.9 & \textbf{85.3} & \textbf{51.2} & \textbf{84.0} & \textbf{11.0} & \textbf{3.2} & \textbf{11.5} &  \textbf{46.7}  &  \textbf{28.4}\\ 
\hline
\end{tabular}
\caption{The performance comparisons among different methods. The best result is highlighted with \textbf{bold} type.}
\label{tbl:accuracy comparisons}
\vspace{-6pt}
\end{table*}

\subsection{Dataset and Metric}
\textbf{Dataset} We conduct the experiments on a public dataset, AVE, which collects the videos from real movie scenes that cover a wide range of genres. A movie scene consists of 35.09 shots on average, and the shot boundaries in seconds have been annotated. Thereby a total number of 196,176 shots are segmented out from all the movie scenes, and the attributes of each shot are manually labeled which helps to evaluate algorithms objectively. We follow the same protocol as AVE, where 70 percent of scenes are used for training, 10 percent for validation, and the left 20 percent are used as the test set. In line with the work, we sample 9 consecutive shots from each scene at a time. The first 4 shots, i.e., n=4, in this sequence are encoded as the initial editing context to predict the attributes of the next shot. For a fair comparison with previous work which formulates the editing process as a retrieval problem, we also perform a retrieve task with the predicted attributes, and the candidate list is composed of the 5 remaining shots.

\noindent \textbf{Evaluation Metrics} For the retrieval task, we evaluate the method with the $rank1$ metric which is defined as the percentage of the correct retrievals where the ground-truth shot is the first shot of the original candidate sequence. For the attribute prediction task, we report average per-class accuracy $Acc_i$ and the overall accuracy $Acc$. Let $\hat{a}_i$ and $a_{i}$ denote the predicted and ground-truth class of the $i$-th attribute, the metrics is defined as:
\begin{align*}
  Acc_i &= \frac{1}{N} \sum^{N} \hat{a}_i == a_{i} \\
  Acc = \frac{1}{N} \sum^{N} (\hat{a}_1 == a_{1})& \&\& \dots \&\& (\hat{a}_8 == a_{8}) = \frac{1}{N} \sum^{N} \hat{A} == A
\end{align*}
where $N$ is the number of samples. The above metrics are designed for one-shot editing, yet editing is a sequential process. To validate the ability to make sequential decisions, we further develop the metrics, two-shot retrieve accuracy $2-rank1$ and two-shot overall attribute accuracy $2-Acc$, to evaluate the two-shot editing ability \cite{Xiong2022}. The two-shot editing metrics are based on the assumption that the decisions for the first shot are completely correct. Let $\hat{S}_{i+n+1}$ and $\hat{S}_{i+n+2}$ denote two consecutive retrieved shots for a sample $\{S_i, \dots, S_{i+n}\}$ , and the corresponding ground-truth shots are $S_{i+n+1}$ and $S_{i+n+2}$, then $2-rank1$ is defined as:
\begin{align*}
  2-rank1 &= \frac{1}{N} \sum^{N} (\hat{S}_{i+n+1} == S_{i+n+1}) \&\& (\hat{S}_{i+n+2} == S_{i+n+2})
\end{align*}
In the calculation of 2-shot accuracy $2-Acc$, the predicted attributes $\hat{A}_{i+n+1}$ and $\hat{A}_{i+n+2}$ are compared with the ground-truth attributes $A_{i+n+1}$ and $A_{i+n+2}$, that is:
\begin{align*}
  2-Acc &= \frac{1}{N} \sum^{N} (\hat{A}_{i+n+1} == A_{i+n+1}) \&\& (\hat{A}_{i+n+2}==A_{i+n+2})
\end{align*}

\begin{table*}[h]
\centering
\begin{tabular}{c|c|c|c|c|c|c|c|c|c|c|c|c|c}
\hline
 Methods&Feat.& $Acc_{np}$ & $Acc_{sa}$ & $Acc_{sl}$ & $Acc_{sm}$ & $Acc_{ss}$ & $Acc_{sub}$ & $Acc_{st}$ & $Acc_{sou}$ & $1-Acc$ & $2-Acc$ &$rank1$ & $2-rank1$ \\
\hline
\multirow{6}{*}{\textbf{RL}} & \textbf{R3D} &  44.8 & 84.2 & 82.7 & 55.3 & 73.0 & 82.1 & 46.4 & 83.2 & 8.4 & 2.3 & 42.4 & 22.5\\ 
& \textbf{CARL}      & 37.6 & 82.4 & 73.1 & 45.2 & 74.9 & 83.1 & 46.1 & 81.0 & 6.9 & 1.8 & 42.6 & 24.0\\
& \textbf{CLIP}      & 38.3 & 83.6 & 74.8 & 51.1 & \underline{75.0} & 83.8 & 47.6 & 83.4 & 7.2 & 1.9 & 42.7 & 23.9\\
& \textbf{XCLIP-vis} & 35.2 & \underline{86.4} & 72.0 & 47.1 & 74.3 & 84.4 & 44.3 & \underline{84.0} & 8.0 & 2.2 & 43.2 & 22.8\\
& \textbf{GT-ad}     & \textbf{59.2} & \textbf{86.5} & \textbf{94.7} & \textbf{73.8} & \textbf{80.1} & \textbf{90.3} & \textbf{59.0} & \textbf{86.4} & \textbf{23.8} & \textbf{10.4} & \textbf{50.5} & \textbf{30.0}\\
\cline{2-14}
& \textbf{XCLIP-ad} & \underline{52.3} & 86.3 & \underline{89.1} & \underline{59.9} & 74.9 & \underline{85.3} & \underline{51.2} & \underline{84.0} & \underline{11.0} & \underline{3.2} & \underline{46.7} & \underline{28.4} \\
\hline
\end{tabular}
\caption{The performance comparisons among the methods using different representations as editing context. Note that \textbf{GT-ad} uses the ground-truth attributes and the second-best results are highlighted with the underline.}
\label{tbl:ablation representation}
\vspace{-6pt}
\end{table*}

\subsection{Implementation Details}
During the representation process in Sec.\ref{sec:context representation}, we adopt the text and video encoders of the pre-trained XCLIP model to extract multi-modal representations, and 32 frames are uniformly sampled from each shot as the input to the video encoder. The editing context length $n$ is set to 4, and the total dimension of attribute distributions is $7+6+3+6+6+9+8+6=51$, thus the dimension of the editing context is $4*51=204$. In our RL-based editing framework, we build the critic network with a Multi-Layer Perception (MLP) network, which takes the input of the concatenation of the editing context and the sampled action. As for the policy network, we adopt the ViT \cite{Dosovitskiy2020} architecture without the patch embedding module, thus the editing context is reshaped as a token sequence with a length of 4 and fed directly to the first transformer block. The critic and policy networks are trained using two Adam optimizers under the Pytorch framework with a learning rate of 1e-4. To speed up the training under the RL framework, we first pre-train a model which has the same architecture as the policy network in a supervised manner, then its parameters are used to initialize the policy network to conduct reinforcement learning.
\subsection{Comparisons}
\label{subsec:comparison}
In this section, we will compare our method with serval baselines in terms of the attribute prediction task. We conduct this experiment with 5 methods: 1) Random, randomly predict the shot attributes; 2) MLP, an MLP network with 8 classification heads is trained with the cross entropy as the objective function, using the features extracted from R3D model; 3)LSTM \cite{Chen2018}, to model the temporal relationship, we implement an LSTM-based network and train it with the R3D features. 4) Trans., the actor network trained in the same way as MLP using the proposed representations. 5) RL, the proposed method. The results are reported in Table \ref{tbl:accuracy comparisons}, it can be observed that the proposed method outperforms all the baseline methods in most of the metrics except for the accuracy of the shot subject attribute $Acc_{ss}$. Especially, for the metrics of $\{1-Acc, 2-Acc\}$, our method improves traditional \textbf{MLP} and \textbf{LSTM} by $\{3.3\%, 0.7\%\}$ and $\{3.1 \%, 0.9\%\}$, respectively. After applying our RL editing method, the reward gained also improves from $11.0$ to $11.5$. These improvements benefit from the proposed representations and RL-based editing framework. Besides attribute prediction, we also compare these methods on the retrieval task. The predicted attributes are used as a query to retrieve the shot with the closest attributes from the candidate list, and the one-shot and two-shot retrieval results, $rank1$ and $2-rank1$, are also listed in Table \ref{tbl:accuracy comparisons}. Even if evaluated on the retrieval task, our method of predicting attributes is still effective and outperforms baseline methods in both metrics consistently.

\begin{figure*}[htbp]
  \includegraphics[scale=0.8]{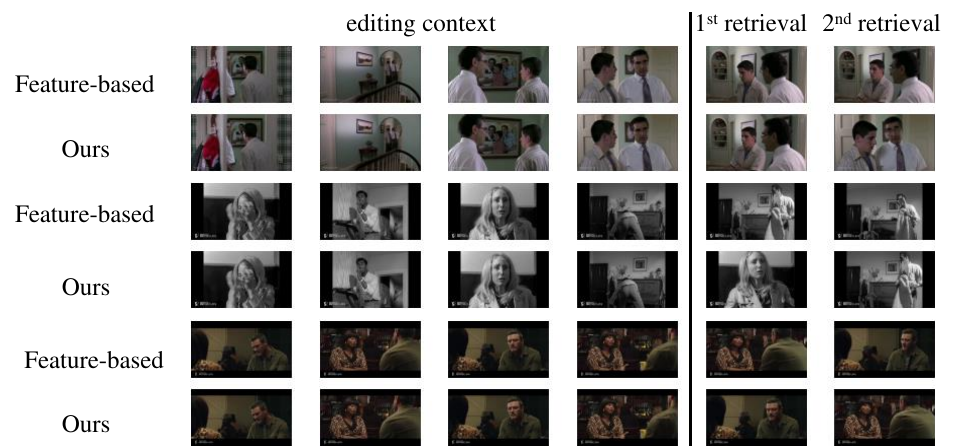}
  \vspace{-6pt}
  \caption{The retrieved sequences generated by our method and a feature-based method.}
  \label{fig:qualitative}
\end{figure*}

\subsection{Ablation Study}
\textbf{Representation} In this section, we will inspect the effectiveness of the proposed context representations. Specifically, we compare the performances of methods using other five kinds of representations, including 1) \textbf{R3D} \cite{Tran2018}, this feature has been used by previous automatic editing systems; 2) \textbf{CARL} \cite{Chen2022}, a self-supervised learning-based video representation model; 3) \textbf{CLIP} \cite{Radford2021}, a cross-modal model trained with image-text pair data. We use the image encoder of CLIP to extract frame-wise features, and the average frame feature is used as the shot representation; 4) \textbf{XCLIP-vis}, XCLIP \cite{Ma2022} is also a cross-modal model trained with video-text pair data. Unlike the proposed representation which calculates the similarities between vision features and textual features, we use the embedding from the video encoder directly as the representation in this experiment; 5) \textbf{GT-ad}, To prove the rationality of the assumption that the attribute information of context shots helps to make correct editing decisions, we encode the ground-truth attributes of context shots as one-hot vectors and cascade these vectors as representations.

The results are listed in Table \ref{tbl:ablation representation}, where \textbf{XCLIP-ad} represents the proposed attribute distribution representation. \textbf{R3D}, \textbf{CARL}, \textbf{CLIP}, and \textbf{XCLIP-vis} are all visual features, so it is not surprising that the methods using them achieve similar performances. However, since they are not designed particularly for the editing task, the performances using them are relatively low when compared with the method using our representation, \textbf{XCLIP-ad}. Specifically, the best feature among them, \textbf{R3D}, attains $2.6\%$ and $0.9 \%$ lower $1-Acc$ and $2-Acc$  than \textbf{XCLIP-ad} dose. Moreover, we can observe that the ground-truth attribute encoding \textbf{GT-ad} outperforms other methods by a large margin and achieves the best result in all metrics. This observation also justifies that the direction of exploring context attributes is promising, even though there is still a gap.

\noindent \textbf{RL-based framework} Another core component in our method is the RL-based editing framework. In this section, we will investigate the benefits of learning to make sequential editing decisions. We apply our RL framework to three baseline methods, i.e., \textbf{MLP}, \textbf{LSTM}, and \textbf{Trans.}, and the results with and without RL are reported in Table \ref{tbl:ablation RL}. By comparing the results in two-shot metrics, i.e., $2-Acc$ and $2-rank1$, where our editing framework consistently improves the baseline methods, it can be concluded that the proposed framework does help to make sequential decisions. Surprisingly, after applying our framework, the performances of baseline methods in one-shot metrics, i.e., $1-Acc$ and $rank1$, are increased as well. The reason is that the first shot decision influences the decision of the second shot, so the framework will adjust the first shot decision to make sequential decisions to achieve higher rewards during the training process.

\begin{table}[h]
\centering
\begin{tabular}{c|c|c|c|c}
\hline
Methods & $1-Acc$ & $2-Acc$ &$rank1$ & $2-rank1$  \\
\hline
\textbf{AVE} \cite{Argaw2022} & - & - & 41.4 & - \\ 
\hline
\textbf{MLP}     & 8.4 & 1.5 & 44.1 & 24.4 \\
\textbf{MLP+RL}  & 9.1 & 2.0 & 45.8 & 25.5 \\
\hline
\textbf{LSTM}    & 9.0 & 2.3 & 45.7 & 25.1 \\
\textbf{LSTM+RL} & 9.9 & 2.5 & 46.0 & 26.0 \\
\hline
\textbf{Trans.}    &  8.4 & 2.4  & 43.7 & 24.7 \\
\textbf{Trans.+RL} &  \textbf{11.0} & \textbf{3.2} & \textbf{46.7} & \textbf{28.4}\\
\hline
\end{tabular}
\caption{The performance comparisons between the methods with (\textbf{+RL}) and without RL.}
\vspace{-24pt}
\label{tbl:ablation RL}
\end{table}

\subsection{Qualitative Results}
Previous method \cite{Argaw2022} has attempted to use the LSTM network to estimate the visual feature to retrieve the next shot, and the features used for training are extracted with a pre-trained R3D model. To intuitively illustrate the benefits of our method, we implement the above method by ourselves and visualize three retrieved sequences from both our attribute-based method and the above feature-based method in Fig.\ref{fig:qualitative}. From this figure, it can be seen that the feature-based method tends to retrieve visually similar shots, e.g., the 1-st row, the 3-rd row, and the 1-st retrieved shot of the fifth row, and these resultant sequences do not follow the empirical editing rules in dialogue scenes \cite{Leake2017}. One reason might be that the shot retrieval is based on the feature similarities between the estimated visual feature and the features of candidate shots, so visually similar shots are preferred by the feature-based method. In contrast, our attribute-based retrieval results in more diverse sequences, which basically follow the shot order in the dialogue scenes, e.g., show the speaking characters alternatively. From this angle, the proposed presentation is more suitable for the editing task.

\begin{figure*}[htbp]
  \includegraphics[scale=0.75]{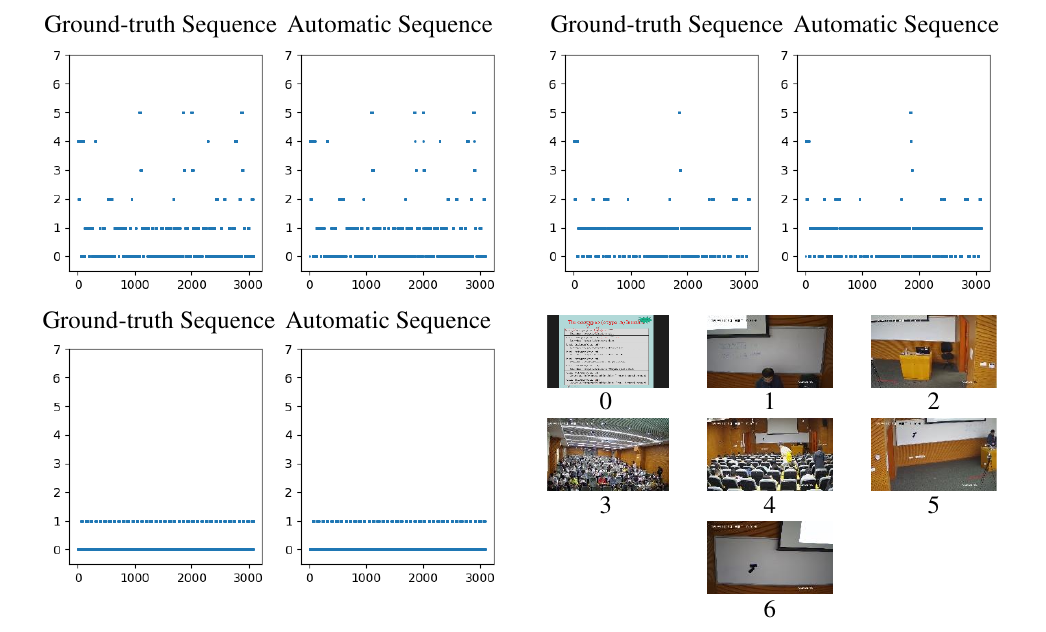}
  \vspace{-12pt}
  \caption{The sequences generated by the trained actors of three styles and the corresponding ground-truth sequences.}
  \label{fig:application}
   \vspace{-12pt}
\end{figure*}

\subsection{Evaluation in Event-Driven Scene}
\textbf{Application background} To verify that our editing framework is also feasible in event-driven directing/editing scenes, we apply our RL-based editing framework to learn the editing/directing style of online lecture broadcasting. The task of automatic online lecture broadcasting is to select a view from multiple cameras to broadcast at a time. A few automatic editing systems \cite{Liu2001,Rui2004,Wang2007} for this task have been studied, yet these heuristic methods only generate a mechanical broadcasting stream and cannot meet the preferences of different students. For example, some students prefer to watch slide view when there is no particular event happening, while some students might focus more on the teacher's view. Therefore, we will conduct experiments to validate that the proposed editing framework can also tackle this problem. In other words, we will show that our method can learn individual preferred styles from some given watching records.

\noindent \textbf{Data preparation} The experiment is conducted in a classroom where 7 cameras are deployed to shoot the class from different angles, including  the slide close-up shot, left and right blackboard close-up shots, left and right medium shots, overall long shot, and student shot. To train and evaluate the policy network, it demands stylistic watching records as the ground-truth selections. As it is challenging to collect the stylistic records, we employ a simulated scheme to generate them. Concretely, we collect 7 synchronized videos with a length of $T$ time units (frame or second) from a real class for training, and 7 videos from another class are used as a testing scene. Next, the representations $F = \{f_t\}_{t=1:T}, f_t = [f^{e}_t|f^{v}_t|f^{s}_t]$ which counts the event information $f^e$, view transition constraints $f^v$, and switch penalty $f^s$, are extracted and fed to a parameterized heuristic editing method $H()$ \cite{Hu2021} to generate a view sequence as watching records. Therefore, the ground-truth view selections, $Y^{tr}=\{y^{tr}_t\}_{t=1:T}$ and $Y^{test}=\{y^{test}_t\}_{t=1:T}$ for training and testing are obtained as :
\begin{align*}
  Y^{tr} &= H(F^{tr}, \omega), \quad Y^{test} = H(F^{test}, \omega)
\end{align*}
where $\omega$ is the parameter vector controlling the styles of view selections. The view sequences generated with different $\omega$ are considered sequences of different styles.

\noindent \textbf{RL problem} Supposing $Y^{tr}$ and $Y^{test}$ are generated with the same $\omega$ on two different classes, the goal of our framework is to train a policy network with data $F^{tr}$ and $Y^{tr}$, which then takes as input $F^{test}$ to generate view sequence approaching $Y^{test}$. The purpose of this experiment is to prove whether the policy network learns the style of $Y^{tr}$ with our framework. As we mainly focus on investigating the learning ability and the feasibility of learning the style from the labeled sequence, the state representation at time $t$ is simply defined as $f_t$, and the action space is defined as the camera indices, $\hat{y}_t \in \{1, \dots, 7\}$. The reward function $R()$ is:
\begin{align*}
  R(\hat{y},y) &=\begin{cases}
                   1, & \mbox{if } \hat{y}==y \\
                   -1, & \mbox{otherwise}.
                 \end{cases}
\end{align*}
when the sampled action $\hat{y}$ is the same as the ground-truth action $y$, the function returns a positive reward, and vice versa. The objective functions for training critic and actor networks remain the same as Eq.\ref{eqn:advatage function} and Eq.\ref{eqn:actor loss}. For simplicity, two MLP models are adopted as the architectures of critic and actor networks.

\noindent \textbf{Evaluation} During the experiment, we use three different $\{\omega_i\}_{i=1:3}$ to generate ground-truth training and testing sequences, $\{Y^{tr}_i\}_{i=1:3}$ and $\{Y^{test}_i\}_{i=1:3}$, and these three paired dataset $\{(Y^{tr}_i,Y^{test}_i)\}_{i=1:3}$ are used to train and evaluate three policy networks separately, representing three virtual editors of different styles. For each style, we report the comparisons between the predicted sequence $\hat{Y}^{test}_i$ and the ground-truth sequence $Y^{test}_i$ in terms of the overlap ratio $Ratio$ and three sequence properties, i.e., the average shot length $L_{avg}$, the maximum shot length $L_{max}$, and the number of switches $N_{sw}$. The results are listed in Table \ref{tbl:stylistic comparisons}, the trained actor can generate sequences that are close to the corresponding ground-truth sequence in testing scenes. For example, the sequence $\hat{Y}^{test}_1$ generated by the actor trained with $Y^{tr}_1$ in the testing class achieves an overlap ratio of $99 \%$ with $Y^{test}_1$, and the properties such as average shot length, the maximum shot length are also close. The same conclusion is also obtained in the other two styles. These comparisons prove that our editing framework is able to learn the editing styles from the stylistic sequences.

In Fig.\ref{fig:application}, we visualize the generated sequences of three stylistic actors and the corresponding ground-truth sequences in a testing class. It is obvious that the editing patterns or styles of the three ground-truth sequences are truly different, e.g., the top-left sequence displays the slide view (0) more frequently, the top-right sequence contains more close-up views (1), and the bottom-left sequence regularly displays the slide view and the close-up view. Thus, the goal of the stylistic actor networks is to generate the corresponding stylistic sequences. The figures of automatic sequences (generated by actor networks) show similar patterns as their ground truth. This result confirms that our RL-based editing framework can learn different editing styles for event-driven scenes.

\begin{table}[h]
\centering
\begin{tabular}{c|c|c|c|c|c|c|c}
\hline
style & $Ratio$ & $L_{avg}$ &$ L_{avg}^{*}$ & $L_{max}$ & $L_{max}^{*}$ & $N_{sw}$ & $N_{sw}^{*}$ \\
\hline
$\omega_1$ &  0.99 & 25.3 & 26.6 & 71 & 71 & 121 & 115  \\ 
$\omega_2$ & 0.98 & 23.3 & 23.7 & 75 & 76 & 131 & 129 \\
$\omega_3$ & 0.99 & 40.0 & 40.0 & 61 & 61 & 76 & 76\\
\hline
\end{tabular}
\caption{The editing results on three stylistic test sets. $*$ indicates the properties of ground-truth sequences. }
\label{tbl:stylistic comparisons}
\vspace{-24pt}
\end{table}

\section{Conclusion}
\label{sec:conclusion}
In this paper, we formulate a new editing task as next shot attribute prediction, which is more helpful in practice compared with retrieval-based editing. Next, a new shot attribute-based editing representation is proposed, experimental results show that this representation benefits general-scene editing and is superior to other visual features. Furthermore, to bridge the gap between the videos generated by heuristic rules and the professional-look videos, we propose an RL-based editing framework to train the virtual editor with professional videos. Extensive experiments are carried out on a real movie dataset to demonstrate that our framework can directly learn the editing patterns from well-edited movies and make sequential editing decisions. Finally, we conduct experiments in an online lecture broadcasting scene, which prove that the RL editing framework can generalize to event-driven editing.

\bibliographystyle{ACM_Reference_Format}
\bibliography{egbib}
\end{document}